\documentclass{article}

\PassOptionsToPackage{numbers, compress}{natbib}

\usepackage[final]{neurips_2023}

\usepackage[utf8]{inputenc} 
\usepackage[T1]{fontenc}    
\usepackage[hidelinks]{hyperref}       
\usepackage{url}            
\usepackage{booktabs}       
\usepackage{amsfonts}       
\usepackage{nicefrac}       
\usepackage{microtype}      
\usepackage{xcolor}         

\usepackage{leftidx}
\usepackage{xcolor}

\usepackage{subfigure}
\usepackage{graphicx}
\usepackage{algorithm} 
\usepackage[noend]{algpseudocode}
\usepackage{mdwtab}
\usepackage{eqparbox}
\usepackage{amsmath,amsfonts,amssymb}
\usepackage[hidelinks]{hyperref}
\hypersetup{colorlinks=true,allcolors=blue}
\usepackage{caption} 
\usepackage{arevmath} 
\usepackage{mathtools}
\usepackage{array}
\usepackage{siunitx}
\usepackage{textgreek}
\usepackage{mathptmx}
\usepackage{comment}
\usepackage{booktabs, multirow} 
\usepackage{soul}
\usepackage{changepage,threeparttable} 

\hypersetup{
    colorlinks=false,
    pdfborder={0 0 0},
    hidelinks
}

\usepackage{tikz}
\usetikzlibrary{shapes.geometric, shapes.multipart, arrows, positioning, calc}
\usepackage{pgfplots}
\pgfplotsset{width=7.3cm,compat=newest}
\newcommand{\ours}{\textsc{Methods}}

\title{The nextAI Solution to the \\ NeurIPS 2023 LLM Efficiency Challenge}

\begin{document}

\author{%
  Gyuwon Park$^{1}$ \quad DongIl Shin$^{2}$ \quad SolGil Oh$^{2}$ \quad SangGi Ryu$^{2}$ \quad Byung-Hak Kim$^{2}$ \\
  $^{1}$Department of Computer Science and Engineering, UNIST \\ $^{2}$CJ Corporation \\
  \texttt{gyuwon12@unist.ac.kr} 
}

\maketitle

\begin{abstract}

The rapid evolution of Large Language Models (LLMs) has significantly impacted the field of natural language processing, but their growing complexity raises concerns about resource usage and transparency. Addressing these challenges, we participated in the \href{https://llm-efficiency-challenge.github.io}{NeurIPS 2023 LLM Efficiency Challenge}, aiming to fine-tune a foundation model within stringent constraints. Our focus was the LLaMa2 70 billion model, optimized on a single A100 40GB GPU within a 24-hour limit. Our methodology hinged on a custom dataset, carefully assembled from diverse open-source resources and benchmark tests, aligned with the challenge’s open-source ethos. Our approach leveraged Quantized-Low Rank Adaptation (QLoRA) Fine tuning, integrated with advanced attention mechanisms like Flash Attention 2. We experimented with various configurations of the LoRA technique, optimizing the balance between computational efficiency and model accuracy. Our fine-tuning strategy was underpinned by the creation and iterative testing of multiple dataset compositions, leading to the selection of a version that demonstrated robust performance across diverse tasks and benchmarks. The culmination of our efforts was an efficiently fine-tuned LLaMa2 70B model that operated within the constraints of a single GPU, showcasing not only a significant reduction in resource utilization but also high accuracy across a range of QA benchmarks. Our study serves as a testament to the feasibility of optimizing large-scale models in resource-constrained environments, emphasizing the potential of LLMs in real-world applications.

\end{abstract}

\section{Introduction}
\label{sec:introduction}

\bibliographystyle{apalike}
The emergence and rapid advancement of LLMs (\citet{openai2023gpt4}, \citet{touvron2023llama-2}) have revolutionized various domains of natural language processing. As these models have grown in complexity and capability, their parameter counts have surged exponentially. Such massive models (\citet{chowdhery2022palm}, \citet{DBLP:GPT3-journals/corr/abs-2005-14165}), although impressive in their performance, come with their share of challenges. First, their ever-increasing size necessitates the use of extensive GPU resources, leading to escalated costs and reduced accessibility for the majority. More recently, there has been a concerning trend where the specifics of the model architectures and their training methodologies are not disclosed, resulting in a lack of transparency and reproducibility.  There is a pressing need for clear and reproducible training procedures, which has been conspicuously absent in many recent developments.

In response to these concerns, the NeurIPS LLM Efficiency Challenge has been introduced. This challenge focuses on adapting a foundation model to specific tasks but with imposed constraints. Participants are required to fine-tune models on a single GPU, either of the 4090 or A100 (40GB) variants, all within a strict 24-hour limit. Such constraints emphasize real-world applicability, simulating scenarios of limited resources. Furthermore, the initiative highlights the significance of open source, promoting not just replicability but also widespread accessibility, thereby eliminating barriers of proprietary restrictions.

In addressing this challenge, our team at nextAI identified a model that thrived under the set constraints. While several models like LLaMA-30b, LLaMA-65b (\citet{touvron2023llama-1}) and Mistral-7b (\citet{jiang2023mistral}) were considered, LLaMA2-13b, LLaMA2-70b stood out, and we fine-tuned it on a single A100 40GB GPU within 24 hours. Key to this achievement were techniques like Quantization, Parameter-Efficient Fine-Tuning (\citet{dettmers2023qlora}, \citet{DBLP:LoRA-journals/corr/abs-2106-09685}, \citet{peft}) and advanced Attention mechanisms (\citet{dao2023flashattention2},\citet{dao2022flashattention}, \citet{DBLP:SWA-journals/corr/abs-2103-14030}, \citet{DBLP:All-you-need-journals/corr/VaswaniSPUJGKP17}). Our custom dataset (\citet{lee2023platypus}, , \citet{zhang2022automatic-cot}, \citet{zhou2023lima}) further optimized the fine-tuning process. This approach emphasized the adaptability of LLMs in constrained environments.

Parameter-Efficient Fine-Tuning (PEFT) (\citet{peft}) enables efficient adaptation of pre-trained language models to downstream applications by fine-tuning a subset of model parameters, reducing computational and storage challenges. Among the available methods, we primarily employed LoRA (Low-Rank Adaptation of Large Language Models) (\citet{DBLP:LoRA-journals/corr/abs-2106-09685}). Thanks to the advancements in this domain, we were able to leverage Quantized-LoRA (QLoRA) (\citet{dettmers2023qlora}), which combined quantization with LoRA, enhancing our model's efficiency. Quantization techniques offer significant benefits by reducing the model size and accelerating inference, making it possible to deploy more complex models on resource-constrained devices with minimal loss of accuracy. Notably, certain QLoRA experiments, such as variations with changing \( \text{lora\_r} \) values and differing LoRA layer targets, lacked exhaustive results. Our team bridged this gap, providing in-depth insights beneficial for subsequent research in the field.

Transformers have come a long way since "Attention is All You Need" (\citet{DBLP:All-you-need-journals/corr/VaswaniSPUJGKP17}) revolutionized the field. As models grow in complexity and the volume of attention computations surges, efficiently managing these computational demands becomes crucial. Our solution is the adoption of the Flash Attention framework (\citet{dao2023flashattention2}, \citet{dao2022flashattention}), which streamlines attention calculations through an innovative IO-aware algorithm, optimizing the synergy between GPU high bandwidth memory (HBM) and on-chip SRAM. This advancement is pivotal for maintaining performance despite the increased complexity of attention mechanisms in larger models. While the Sliding Window Attention (SWA) (\citet{DBLP:SWA-journals/corr/abs-2103-14030}) mechanism also contributes to computational efficiency, it's the incorporation of Flash Attention that stands out as a transformative measure to sustain model scalability and computational tractability.

Our training strategy hinged on the exclusive use of open-source data, assembling a corpus that aligns with our model's training needs. Influenced by \citet{chung2022scaling-diverse}, we crafted a dataset with a spectrum of instructions for robust context coverage. Insights from top performers on the HuggingFace Open LLM Leaderboard refined our dataset design. The composition details are in Table 1. We adopted the Alpaca format (\citet{alpaca}) for instruction tuning, leveraging its effectiveness in task comprehension and execution. This dual approach of dataset diversity and instructional clarity is the foundation for our ensuing performance analysis.

\section{Related Work}
\label{sec:relatedwork}

\subsection{Parameter-Efficient Fine-Tuning}
In the evolving field of LLMs, a suite of PEFT techniques has significantly diminished the necessity for comprehensive model fine-tuning. Among these, LoRA (\citet{DBLP:LoRA-journals/corr/abs-2106-09685}) stands out by adapting models through low-rank matrix updates, while methods like Prefix Tuning (\citet{DBLP:Prefix-journals/corr/abs-2101-00190}) enhance model performance by optimizing continuous prompts. Meanwhile, P-Tuning (\citet{DBLP:P-tuning-journals/corr/abs-2110-07602}) illustrates the untapped potential of LLMs to comprehend complex contexts with minimal parameter adjustments. Other methods, including Prompt Tuning (\citet{DBLP:Prompt-journals/corr/abs-2104-08691}) and IA3 (\citet{liu2022fewshot}), focus on refining prompts and modulating inner activations, respectively. 

The publication "Towards A Unified View of Transfer Learning" (\citet{DBLP:Towards-journals/corr/abs-2110-04366}) offers a substantial contribution to the field of transfer learning for LLMs. It introduces a comprehensive framework that reframes existing approaches as precise modifications to the hidden states of pretrained models. This framework outlines various design dimensions, such as the specific functions for computing modifications and their points of application, facilitating more sophisticated and effective transfer learning tactics. This integrated approach offers a nuanced perspective on tailoring PEFT methods to leverage the innate strengths of LLMs.

\subsection{Quantization and LoRA}
In tandem with the growth of LLMs, quantization has emerged as a pivotal compression technique, enhancing efficiency and scalability by reducing the bit width of parameters and activations (\citet{dettmers2022llmint8}, \citet{dettmers2023qlora}). Efforts have primarily centered on maintaining the fidelity of LLMs post-quantization, with a focus on minimizing the computational load without necessitating retraining. Innovative strides in the field of LLM quantization have been made to mitigate the impact of outliers in parameter distributions, fueling the development of both adaptive and dynamic quantization techniques, as well as sophisticated clustering methods that apply customized quantization to various parameter sets. Alongside these strides, the integration of LoRA with quantization techniques has been a game-changer, notably with QLoRA demonstrating marked efficiency gains. Building upon this, QA-LoRA (\citet{xu2023qalora}) has accelerated processing speeds. Meanwhile, LongLoRA (\citet{longlora}) has effectively addressed the efficiency issues encountered in fine-tuning with longer contexts.

\subsection{Attention and Memory}
Transformers (\citet{DBLP:All-you-need-journals/corr/VaswaniSPUJGKP17}) struggle with the computational intensity of self-attention for long sequences, but Flash Attention (\citet{dao2022flashattention}) mitigates this by optimizing memory exchanges within GPU architecture, significantly enhancing efficiency. The subsequent Flash Attention version 2 (\citet{dao2023flashattention2}) further refines this approach, with better work partitioning and parallelization strategies, halving the computational demands and nearing the efficiency of optimized matrix multiplications, a step eagerly adopted by the open-source community.

Advancing beyond Flash Attention's efficiencies, innovative methods like LLaMA2's Grouped-query attention (GQA) (\cite{touvron2023llama-2}) and Mistral's (\citet{jiang2023mistral}) Sliding Window Attention (SWA) (\citet{DBLP:SWA-journals/corr/abs-2103-14030}) have emerged, offering faster inference and managing longer sequences with reduced computational load. Additionally, LongLoRA's Shift Short Attention (\citet{longlora}) has optimized attention computation through intelligent structuring. Together, these advancements represent substantial strides toward minimizing GPU memory usage and shortening computational duration.

\subsection{Instruction Tuning}
Instruction tuning in contemporary research broadly divides into two primary strategies. The first leverages data integration from annotated natural language datasets, exemplified by the likes of the FLAN (\citet{DBLP:FLAN-journals/corr/abs-2109-01652}), where text-label pairs are ingeniously transformed into (instruction, output) pairs via templates. The second paradigm shifts towards generating datasets with the aid of LLMs such as GPT-3.5-Turbo or GPT-4 (\citet{openai2023gpt4}). This method capitalizes on the efficiency of LLMs to produce outputs from both manually curated and LLM-expanded instructions, giving rise to datasets such as InstructionWild (\citet{instructionwild}) and Self-Instruct (\citet{selfinstruct}). Building on these foundational approaches, the field has seen novel data formats such as the Stanford Alpaca (\citet{alpaca}), complementing projects like LLaMA, and datasets akin to Orca (\citet{mukherjee2023orca}), which re-envision FLAN through the capabilities of LLMs. Amidst these advancements, the LIMA (\citet{zhou2023lima}) study emerged, emphasizing the importance of high-quality datasets, inspiring the Open-Platypus (\citet{lee2023platypus}) dataset which synthesizes these varied approaches. This dataset, noted for its high caliber, has become an almost indispensable resource in the STEM fields and consistently ranks highly on the Hugging Face open LLM leaderboard, signaling its critical role in current research and applications.

These research endeavors and the inherent challenges they address underscore a concerted effort to craft high-quality datasets, reflecting the field's commitment to precision and robustness in data-driven models.

\begin{table}[t]
\centering
\begin{tabular}{|p{4.5cm}|p{3.5cm}|p{1cm}|p{3cm}|}
\hline
\textbf{Dataset Name} & \textbf{Task} & \textbf{\# Size} & \textbf{License Type} \\ \hline
PRM800K: A Process Supervision Dataset \cite{lightman2023lets-prm800k} & Math QA & 13k & MIT \\ \hline
ScienceQA: Science Question Answering \cite{lu2022learn-ScienceQA} & Science QA & 1.3k & Creative Commons Attribution-NonCommercial-ShareAlike 4.0 \\ \hline
ReClor: A Reading Comprehension Dataset Requiring Logical Reasoning \cite{DBLP:ReClor-journals/corr/abs-2002-04326} & Logical reasoning QA & 4.5k & Non-commercial \\ \hline
TheoremQA: A Theorem-driven Question Answering Dataset \cite{chen2023theoremqa} & STEM QA & 0.5k & MIT \\ \hline
tigerbot-kaggle-leetcodesolutions-en-2k \cite{tigerresearch2023tigerbot} & Coding problem solving & 0.4k & Apache-2.0 \\ \hline
ARB: Advanced Reasoning Benchmark for Large Language Models \cite{sawada2023arb} & STEM QA & 0.7k & MIT \\ \hline
Openassistant-guanaco \cite{dettmers2023qlora} & Conversation & 0.8k & Apache-2.0 \\ \hline
Multi-News \cite{alex2019multinews} & Summarization & 4k & Other \\ \hline
FaithDial \cite{dziri2022faithdial} & Multi-turn conversations & 4.5k & MIT \\ \hline
LIMA: Less Is More for Alignment \cite{zhou2023lima} & Conversations & 1k & CC BY-NC-SA \\ \hline
Codegen \cite{codegen} & Coding problem solving & 4k & ~ \\ \hline
Dolly \cite{DatabricksBlog2023DollyV2} & Creative Writing, Closed QA, Open QA, Summarization, Information Extraction, Classification, Brainstorming & 15k & cc-by-sa-3.0 \\ \hline
\end{tabular}
\caption{Datasets, Tasks, Number of Data Size, and Licenses. Our datasets, primarily showcased under Open-Platypus \cite{lee2023platypus}, demonstrate a strong emphasis on STEM and logical reasoning. In addition to these specialized datasets, we have also incorporated a diverse range of tasks such as general QA and conversation. This approach is part of our strategic effort to conduct instruction tuning across various fields, ensuring a comprehensive and robust training for diverse applications.}
\label{tab:my_label}
\end{table}

\section{\ours~}
\label{sec:method}


In this section, we delineate the methodologies employed in our challenge submission. Adhering to the guidelines of the challenge and taking cues from the current trends in benchmark testing and recent scholarly advancements, we crafted a bespoke dataset tailored to align with the specificities of the task. This dataset was meticulously designed to resonate with the LLaMa2 70B model, forming the bedrock of our training process. Furthermore, we harnessed the power of QLoRA Fine-tuning, a cutting-edge technique, coupled with advanced attention mechanisms, to optimize the model’s performance. These collective methodologies embody our strategic approach to addressing the challenge’s complexities.

\subsection{Custom Dataset}

The foundation of our methodology is predicated on the creation of a custom dataset, meticulously aligned with the challenge's emphasis on open-source materials and benchmark tests encompassing logic reasoning types of multiple-choice QA scenarios and conversational chat tasks. This alignment necessitated the selection and composition of a dataset that adheres strictly to open-source principles, specifically excluding any datasets generated by LLMs themselves due to the challenge's guidelines.

We used the Open-Platypus\cite{lee2023platypus}  dataset as the reference dataset. Our choice of the Open-Platypus was driven by its reputation as a high-quality, open-source dataset in the STEM field, as evidenced by its frequent use among top performers on the HuggingFace Open LLM Leaderboard. In line with the recommendations from the \cite{zhou2023lima}, our focus was on maintaining high quality within custom dataset. In this case, we used the method utilized in the platypus's data curation. We utilized Sentence Transformer's embeddings (\citet{DBLP:Sentence-journals/corr/abs-1908-10084}) to assess the cosine similarity between sentences, aiming to curate a dataset of the highest calibre. To compensate for the absence of approximately 4k coding-related data excluded due to LLM origination, we integrated selections from the codeGen\cite{codegen} dataset, ensuring diversity and richness in our data.

In our quest to fine-tune the model effectively within the 24-hour constraint, we also incorporated datasets covering diverse tasks. Drawing inspiration from \cite{chung2022scaling-diverse}, we observed that diverse instruction tuning significantly enhances performance. We began customizing with the Alpaca format used in Platypus as a base, carefully adapting it to suit our specific requirements. The data composition underwent rigorous testing across eight different configurations, incorporating additional datasets like CoT\cite{zhang2022automatic-cot} to experiment with various permutations. Ultimately, we established our dataset structure as outlined in \textbf{Table 1}. Furthermore, we implemented a strategy to identify and eliminate any training data with a cosine similarity exceeding 0.9, ensuring the uniqueness and quality of our dataset. The experimental results of these eight data compositions can be found in \textbf{Section 4 Experiment}.

This comprehensive approach to dataset creation and optimization not only adhered to the challenge's stipulations but also positioned us advantageously to leverage the full potential of our selected LLM, laying a solid foundation for the fine-tuning process that follows.

\subsection{QLoRA Fine-tuning}

In fine-tuning, our team utilized the huggingface's \cite{peft} library, further optimized and extended in custom-developed Axolotl library \cite{axolotl}. This enhanced library integrates a wide range of advanced functionalities such as LoRA, QLoRA, flash attention, xformers attention, and gptq, ensuring a comprehensive and versatile toolkit tailored to our specific requirements. Specifically, for this study, our team applied QLoRA technique in conjunction with flash attention. This combination was chosen due to the efficiency and effectiveness of flash attention in handling large-scale language models, particularly in contexts demanding high computational efficiency and low memory footprint.

\textbf{Low-Rank Adaptation} The concept of low-rank adaptation, introduced by \citet{DBLP:LoRA-journals/corr/abs-2106-09685}, enables these models to efficiently adapt within a reduced subspace. We apply this concept by reparametrizing the update to the pre-trained weight matrix \( W_0 \in \mathbb{R}^{d \times k} \) as a low-rank decomposition: \( W_0 + \Delta W = W_0 + BA \), where \( B \in \mathbb{R}^{d \times r} \), \( A \in \mathbb{R}^{r \times k} \), and the rank \( r \ll \min(d, k) \). In this structure, \( W_0 \) remains static during training, while \( A \) and \( B \) are trainable, representing the adaptation.

For an input \( x \), the layer's output in the original network is \( h = W_0x \). With low-rank adaptation, the forward pass is expressed as:

\begin{equation}
 h = W_0x + \Delta Wx = W_0x + BAx 
\end{equation}

This reparametrization begins with \( A \) having a random Gaussian initialization and \( B \) set to zero, ensuring \( \Delta W = BA \) initially equals zero. To ensure balanced learning, \( \Delta Wx \) is scaled by \( \alpha r \), with \( \alpha \) being a constant proportional to \( r \). This scaling negates the need for extensive hyperparameter retuning when \( r \) changes, facilitating a more streamlined adaptation process.

\textbf{Quantization} The adoption of quantization techniques was crucial for deploying the LLaMa2 70B model on a single A100 40GB GPU. While various quantization methods exist, ranging from 8-bit to 2-bit, we placed our trust in the QLoRA method, primarily due to its strong performance as reported in \cite{dettmers2023qlora}. In our implementation, the following QLoRA configuration was used:

\begin{equation}
Y_{\text{BF16}} = X^{\text{BF16}} \text{doubleDequant}(c_{\text{1}}^{\text{FP32}}, c_{\text{2}}^{k-\text{bit}}, W^{\text{NF4}}) + X^{\text{BF16}} L_{\text{1}}^{\text{BF16}} L_{\text{2}}^{\text{BF16}},
\end{equation}

where \textbf{doubleDequant(·)} is defined as:

\begin{equation}
\text{doubleDequant}(c_{\text{1}}^{\text{FP32}}, c_{\text{2}}^{k-\text{bit}}, W^{k-\text{bit}}) = \text{dequant}(\text{dequant}(c_{\text{1}}^{\text{FP32}}, c_{\text{2}}^{k-\text{bit}}), W^{4\text{bit}}) = W^{\text{BF16}}.
\end{equation}

For parameter updates, gradients are computed only for LoRA parameters using 16-bit BrainFloat. This quantization configuration is detailed in \textbf{Table 2}. The equation (1) is introduced in the \cite{DBLP:LoRA-journals/corr/abs-2106-09685}, while equations (2) and (3) are adopted from the \cite{dettmers2023qlora}.

\textbf{QLoRA Configuration}:
\begin{itemize}
    \item \( Y_{\text{BF16}} \): The output tensor in BF16 format, where BF16 refers to 16-bit BrainFloat representation.
    \item \( X_{\text{BF16}} \): The input tensor, also in BF16 format.
    \item \( c^{\text{FP32}}_{1} \), \( c^{k-\text{bit}}_{2} \): Quantization constants used in the double quantization process. Notably, \( c_{2} \) uses the FP8 format.
    \item \( L_{\text{1}}^{\text{BF16}} \), \( L_{\text{2}}^{\text{BF16}} \): Coefficients for additional linear transformations, represented in BF16 format.
\end{itemize}

\begin{table}[ht]
\centering
\begin{tabular}{ll}
\hline
\textbf{Configuration Parameter} & \textbf{Value}\\ 
\hline
quantization method & bitsandbytes \\
load in 8bit & False \\
load in 4bit & True \\
llm int8 threshold & 6.0 \\
llm int8 skip modules & None \\
llm int8 enable fp32 cpu offload & False \\
llm int8 has fp16 weight & False \\
bnb 4bit quantization type & nf4 \\
bnb 4bit use double quantization & True \\
bnb 4bit compute dtype & bfloat16 \\
\hline
\end{tabular}
\caption{Bitsandbytes Quantization Config and Value.}
\end{table}

\textbf{Hyperparameters for QLoRA} In setting up our model, taking cues from the Platypus \cite{lee2023platypus} approach, we determined our learning rate and LoRA α values accordingly in \textbf{Table 3}. In our study, we conducted a hyperparameter search as introduced in the \cite{dettmers2023qlora}. This search for LoRA included varying parameters such as LoRA dropout, LoRA rank with values in {4, 8, 16, 32, 64}, and the configuration of LoRA layers, which included options like key+query, all attention layers, all FFN layers, all layers, and attention + FFN output layers. Our focus was on optimizing LoRA r values and layer combinations to maximize performance within limited resources. According to \cite{dettmers2023qlora}, LoRA r is unrelated to final performance if LoRA is used on all layers. we managed to reduce GPU usage by lowering the LoRA r values and discovered that targeting attention + FFN output layers, as opposed to all layers, maintained comparable performance levels. Additionally, to further minimize GPU consumption, we employed strategies such as using low bit optimizers, adjusting sequence lengths, and reducing batch sizes. While these approaches initially posed challenges in training speed, leveraging Flash attention 2 significantly improved this aspect.

\begin{table}[ht]
\centering
\begin{tabular}{ll}
\hline
\textbf{Hyperparameters} & \textbf{Llama2-70B}\\ 
\hline
batch size & 4 \\ 
micro batch size & 1 \\ 
num epochs & 1 \\ 
learning rate & 3e-4 \\ 
cutoff len & 1024 \\ 
lora r & 4 \\ 
lora $\alpha$ & 16 \\
lora dropout & 0.05 \\ 
lora target modules & attention + FFN output layer \\ 
train on inputs & False \\ 
add eos token & True \\ 
group by length & True \\ 
prompt template & alpaca \\ 
optimizer & paged adamw 8bit \\
lr scheduler & cosine \\ 
weight decay & 0 \\
warmup steps & 100 \\ 
\hline
\end{tabular}
\caption{Hyperparameters for QLoRA with 70B Models.}
\end{table}

Through these methods, we successfully fine-tuned the LLaMa2 70B model on a custom-dataset within \textbf{16 hours}, utilizing \textbf{a single A100 GPU with 39.56GB usage}.

\section{Experiment}
\label{sec:expresults}

Our experiments were conducted in two major phases. Initially, we focused on fine-tuning large-scale language models on a single A100 40GB GPU, emphasizing efficient resource utilization. Following this, we delved into identifying the optimal dataset composition within these constraints, ensuring the most effective training under limited computational resources.

\textbf{Models} We used a range of models, including 30B and 65B LLaMA1, 13B and 70B LLaMA2 models, and the Mistral 7B. The majority of our GPU usage tests were primarily conducted on the 70B models. Upon discovering the feasibility of fine-tuning the 70B model within our resource limits, we embarked on tuning this model with eight different versions of datasets. In our comparative analysis of the final dataset versions across these five models, the 70B model consistently outperformed others in terms of performance. Consequently, we decided to use the 70B model for our final implementation.

\textbf{GPU Usage Tests} For our GPU usage experiments, we focused on optimizing the hyperparameters based on the default QLoRA configuration provided by the Axolotl library \cite{axolotl}. To expedite GPU testing, we utilized the alpaca 2k test  dataset \cite{alpaca2k} for comparing loss metrics. This approach allowed us to quickly evaluate and refine our settings, ensuring optimal performance within our computational constraints. In section 4.2, you can compare detailed experimental results.

\textbf{Benchmark Tests} For our benchmark testing, we employed the HELM \cite{liang2023holistic} system, focusing on a subset of the Massive Multitask Language Understanding (MMLU \cite{DBLP:MMLU-journals/corr/abs-2009-03300}), which includes 17 out of the 57 subjects, as well as BBQ \cite{DBLP:BBQ-journals/corr/abs-2110-08193}, TruthfulQA \cite{DBLP:TruthfulQA-journals/corr/abs-2109-07958}, and CNN/DailyMail \cite{DBLP:CNN-DM-conf/nips/HermannKGEKSB15}. Each test consisted of 50 examples, and the decoding system adhered to the standard settings provided by the HELM test environment. This setup was in alignment with the challenge leaderboard evaluation system. Based on these outcomes, we finalized the model configuration for our challenge submission.

\subsection{Resource Efficiency Optimization}

\textbf{LoRA Configuration} We extensively explored the configuration of LoRA to optimize the model's performance while managing resource constraints. Keeping the LoRA $\alpha$ value fixed at 16, we investigated the impact of varying the rank parameter LoRA r. As illustrated in Table 4, changing the r value showed a clear correlation with GPU usage and training loss, highlighting the delicate balance between computational efficiency and model accuracy. Specifically, we found that reducing the r value from 64 to 4 led to a modest decrease in GPU usage (from 51.27 to 46.49), with only marginal differences in training loss. Consequently, for our fine-tuning process, we settled on a r value of 4, optimizing both resource usage and model performance.

Further investigations into LoRA target layer combinations, as shown in Table 5, revealed that targeting specific layers for adaptation significantly impacts GPU utilization, training loss, and overall training time. These layer combinations lead to reduced GPU usage and shorter training times, with the increase in training loss being negligible compared to the baseline (all layers). Notably, configurations like `attention + FFN output layers` demonstrated optimal balances between GPU usage and training efficiency, indicating their suitability for resource-limited settings.

\begin{table}[ht]
\centering
\begin{tabular}{lllll}
\hline
\textbf{LoRA r} & 4 & 8 & 16 & 64\\ 
\hline
GPU Usage & 46.49 & 46.95 & 47.43 & 51.27 \\
Train loss & 0.822 & 0.824 & 0.825 & 0.825 \\
\hline
\end{tabular}
\caption{Impact of LoRA r Value on GPU Usage and Training Metrics. Unlike the hyperparameters in Table 2, this table reflects a scenario where the cutoff length is set to 4096 and the optimizer used is paged adamw 32bit.}
\end{table}

\begin{table}[ht]
\centering
\begin{tabular}{llll}
\hline
\textbf{LoRA Layer Combination} & GPU Usage & Train loss & Time\\ 
\hline
key+query & \textbf{41.39} & 0.830 & \textbf{22m 12s} \\
all attention layers & \textbf{41.57} & 0.801 & \textbf{22m 43s} \\
all FFN layers & 41.72 & \textbf{0.798} & 23m 30s \\
all layers & 42.09 & \textbf{0.794} & 24m 4s \\
attention + FFN output layers & \textbf{41.46} & \textbf{0.798} & \textbf{22m 30s} \\
\hline
\end{tabular}
\caption{Impact of LoRA Layer Combinations on Training Metrics. The combinations of layers in this table were inspired by and executed based on methodologies from the \cite{dettmers2023qlora}. This approach aimed at maximizing efficiency without compromising performance, allowing for a nuanced comparison of different layer combinations and their impact on training metrics.}
\end{table}

\textbf{Flash Attention 2} The integration of Flash Attention 2 in our QLoRA setup significantly influenced our model's computational performance and resource utilization, as Table 6 clearly demonstrates. When compared with the QLoRA-only approach, the integration of Flash Attention 2 resulted in a substantial reduction in GPU usage from 68.13 to 46.95 and nearly halved the training time from 36m 12s to 19m 18s, with only a negligible increase in training loss. It's to note that the experiment without Flash Attention 2 was unique to this particular setup, providing a clear contrast in performance metrics.

\begin{table}[ht]
\centering
\begin{tabular}{lll}
\hline
\textbf{Method} & QLoRA + Flash Attention 2 & QLoRA Only \\ 
\hline
GPU Usage & 46.95 & 68.13 \\
Train loss & 0.824 & 0.820 \\
Time & 19m 18s & 36m 12s \\
\hline
\end{tabular}
\caption{Comparison of QLoRA with and without Flash Attention 2. This experiment was conducted under the conditions where LoRA r value was set at 8, as referenced in Table 4.}
\end{table}

\subsection{Data Composition for Instruction Tuning}
\textbf{Dataset Composition} Regarding our strategy for dataset composition, we meticulously tested and analyzed the performance of different dataset combinations across various model versions, as detailed in Table 7. These combinations included various proportions of datasets like Dolly, Filtering Platypus, Codegen, FaithDial, Multi-News, Lima, and FLAN-v2-Cot. Filtering platypus refers to a dataset other than the one created by LLM in compliance with the challenge. A notable aspect of our experiment was the comparison between Version 2 and Version 3, which were identical in dataset composition but differed only in the number of training epochs. This contrast was implemented to examine the impact of increasing epochs from 1 to 2 on model performance.

\begin{table}[ht]
\centering
\begin{tabular}{|l|p{11cm}|}
\hline
 & \textbf{Dataset Combination} \\ 
\hline
Version 1 & Filtering Platypus \\
Version 2 & Dolly(100\%) + Filtering Platypus + Codegen(10\%) + Faithdial(20\%) + Multi-News(10\%) \\
Version 3 & Dolly(100\%) + Filtering Platypus + Codegen(10\%) + Faithdial(20\%) + Multi-News(10\%) \\
Version 4 & Dolly(100\%) + Filtering Platypus + Codegen(10\%) + Faithdial(40\%) + Multi-News(20\%) \\
Version 5 & Dolly(100\%) + Filtering Platypus + Codegen(10\%) + Faithdial(20\%) + Multi-News(10\%) + Lima(100\%) \\
Version 6 & Dolly(100\%) + Filtering Platypus + Codegen(10\%) + Lima(100\%) \\
Version 7 & Dolly(100\%) + Filtering Platypus + Codegen(10\%) + Lima(100\%) + FLAN-v2-Cot(2k) \\
\hline
\end{tabular}
\caption{Model Version of Dataset Combination. 'Filtering Platypus' mentioned in each version represents a combined dataset, which includes all data from rows 1 to 7 as listed in Table 1.}
\end{table}

Our benchmark tests, summarized in Table 8, utilized a variety of datasets: MMLU (comprising 17 subjects), BBQ, TruthfulQA, and CNN/DailyMail. For CNN/DailyMail, we recorded five distinct metrics, listed in sequence. The results from these benchmarks were intriguing, revealing that Version 5, despite its intricate dataset combination, emerged as the most effective version. Notably, Version 5 not only showed robust performance across all metrics but also demonstrated a significant balance in performance across different types of tasks.

In the context of the challenge, scores were computed using a mean-win-rate \cite{perlitz2023mean-win} calculation, which essentially quantifies the frequency of a model outperforming others across various comparison scores. Based on this scoring methodology, our Model Version 5 achieved the highest results on the leaderboard in our models. Consequently, we selected Version 5 for our final submission, underlining its superior performance in balancing resource efficiency with high accuracy across a diverse range of benchmarks.

\begin{table}[ht]
\centering
\begin{tabular}{lllll}
\hline
\textbf{Model} & MMLU  & BBQ & TruthfulQA & CNN/DailyMail \\ 
\hline
Version 1 & 0.70 & 0.92 & 0.74 & 0.14, 0.67, 0.42, 0.38, 0.31 \\
Version 2 & 0.71 & 0.92 & 0.76 & 0.16, 0.67, 0.45, 0.67, 0.19 \\
Version 3 & 0.71 & 0.96 & 0.54 & 0.15, 0.67, 0.45, 0.44, 0.15 \\
Version 4 & 0.72 & 0.84 & 0.72 & 0.17, 0.67, 0.45, 0.52, 0.19 \\
\textbf{Version 5} & 0.73 & 0.90 & 0.76 & 0.17, 0.67, 0.44, 0.42, 0.17 \\
Version 6 & 0.69 & 0.96 & 0.7 & 0.15, 0.56, 0.36, 0.44, 0.22 \\
Version 7 & 0.70 & 0.84 & 0.64 & 0.14, 0.67, 0.5, 0.44, 0.15 \\
\hline
\end{tabular}
\caption{Result of Benchmark Test.The testing environment for these benchmarks was identical to that used in the challenge leaderboards. For the MMLU scores, only 17 out of 57 categories were utilized. Additionally, all datasets were evaluated based on 50 examples each to calculate the values. The columns under CNN/DailyMail represent ROUGE-2, Stereotypes (race), Stereotypes (gender), Representation (race), and Representation (gender), respectively.}
\end{table}

\textbf{Concluding Observations} Our comprehensive experimentation culminated in achieving an efficient fine-tuning process for the LLaMA2 70B model. Utilizing Dataset Version 5, we successfully optimized the model to operate within the constraints of a single GPU, showing a total GPU usage of only 39.56 GB. Remarkably, we managed to train the model with a mere 0.0247\% of trainable parameters, completing the process in just 17 hours. This result highlights the effectiveness of our fine-tuning strategy in balancing computational resources and model performance.

\begin{table}[ht]
\centering
\begin{tabular}{lllll}
\hline
\textbf{Model} & Dataset & GPU Usage  & Trainable Parameter & Training Time \\ 
\hline
LLaMA2 70B & Version 5 & 39.56 & 0.0247\% (17M) & 17h \\
\hline
\end{tabular}
\caption{Final Result. This table represents our final training strategy. The actual number of trainable parameters involved in this setup is approximately 17 million.}
\end{table}
\section{Conclusion}
\label{sec:conclusion}
\textbf{Efficiency} Our team's innovative approach to fine-tuning the Llama 2 70B parameter model has demonstrated remarkable efficiency. Leveraging the power of a single A100 GPU with 40GB of GPU memory, we successfully completed the task within a 24-hour (1-day) time frame. This feat was achieved through the strategic application of several advanced techniques, including QLoRA, Flash Attention 2, and multiple dataset curation. Given the same amount of train data volume, this task would typically require 8 A100 GPUs with 80GB. This needs a much longer time frame if none of these strategies were used.

\textbf{Performance} In the challenge ranking, our performance was evaluated through a two-stage process: open evaluation and secret evaluation. The evaluations proceeded in the order of open followed by secret, each utilizing distinct datasets. In the open evaluation, among 33 participating teams, our team secured the 7th place. The secret evaluation was conducted only for the top half of the teams from the open evaluation, where we competed among 17 teams and achieved the 11th rank. 

While we demonstrated strong performance in QA task during the open evaluation, our ranking in the secret evaluation was affected by some unforeseen issues. Notably, a few metrics unexpectedly recorded 'NULL' values, leading to a lower ranking in the secret evaluation. This discrepancy in the results suggests areas for further investigation and improvement in our model's robustness across diverse datasets.

\textbf{Final Remarks} This project underscores the potential of resource-efficient methodologies in achieving high-quality results in machine learning tasks. Our findings highlight the importance of innovative techniques like QLoRA and Flash Attention 2 in optimizing the use of computational resources while maintaining the performance. Moving forward, our focus will be on refining these methods to bolster our model's performance across a wider range of datasets and metrics, paving the way for more accessible and robust AI solutions in the future.

\section*{Acknowledgements}

We would like to express our gratitude to the following individuals for their invaluable contributions to this research. We thank Park, Se Un for his thorough and insightful review of this technical report, which significantly enhanced the quality and clarity of our work.

\bibliography{neurips_2023}

\appendix
\newpage

\renewcommand{\thepage}{}
\setcounter{figure}{0}
\renewcommand{\thefigure}{A\arabic{figure}}
\setcounter{table}{0}
\renewcommand{\thetable}{A\arabic{table}}

\renewcommand{\thesection}{A}

\end{document}